\documentclass[letterpaper]{article} 
\usepackage{aaai25}  
\usepackage{times}  
\usepackage{helvet}  
\usepackage{courier}  
\usepackage[hyphens]{url}  
\usepackage{graphicx} 
\usepackage{natbib}  
\usepackage{caption} 
\usepackage{algorithm}
\usepackage{algorithmic}
\usepackage{amssymb}
\usepackage{amsmath}
\usepackage{multirow}
\usepackage{multicol}
\usepackage{array}
\usepackage{tabularx}
\usepackage{graphicx}
\usepackage{booktabs}
\usepackage{caption}
\usepackage{subcaption}
\usepackage[usenames,dvipsnames]{color}
\usepackage{colortbl}
\definecolor{mygray}{gray}{.9}
\usepackage{color}
\usepackage{newfloat}
\usepackage{listings}
\DeclareCaptionStyle{ruled}{labelfont=normalfont,labelsep=colon,strut=off} 
\floatstyle{ruled}
\newfloat{listing}{tb}{lst}{}
\floatname{listing}{Listing}
\title{TextRefiner: Internal Visual Feature as Efficient Refiner \\ for Vision-Language Models Prompt Tuning}
\author {
    Jingjing Xie\textsuperscript{\rm 1},
    Yuxin Zhang\textsuperscript{\rm 1},
    Jun Peng\textsuperscript{\rm 1},
    Zhaohong Huang\textsuperscript{\rm 1},
    Liujuan Cao\textsuperscript{\rm 1 \footnote{Corresponding author}}
}
\affiliations {
    \textsuperscript{\rm 1}  Key Laboratory of Multimedia Trusted Perception and Efficient Computing, Ministry of Education of China,\\ Xiamen University, Xiamen, China

}

\usepackage{bibentry}

\begin{document}

\maketitle
\begin{abstract}
{
Despite the efficiency of prompt learning in transferring vision-language models (VLMs) to downstream tasks, existing methods mainly learn the prompts in a coarse-grained manner where the learned prompt vectors are shared across all categories. 
Consequently, the tailored prompts often fail to discern class-specific visual concepts, thereby hindering the transferred performance for classes that share similar or complex visual attributes.
Recent advances mitigate this challenge by leveraging external knowledge from Large Language Models (LLMs) to furnish class descriptions, yet incurring notable inference costs.
In this paper, we introduce TextRefiner, a plug-and-play method to refine the text prompts of existing methods by leveraging the internal knowledge of VLMs.
Particularly, TextRefiner builds a novel local cache module to encapsulate fine-grained visual concepts derived from local
tokens within the image branch.
By aggregating and aligning the cached visual descriptions with the original output of the text branch, TextRefiner can efficiently refine and enrich the learned prompts from existing methods without relying on any external expertise.
For example, it improves the performance of CoOp from 71.66 \% to 76.94 \% on 11 benchmarks, surpassing CoCoOp which introduces instance-wise features for text prompts. Equipped with TextRefiner, PromptKD achieves state-of-the-art performance and is efficient in inference. Our code is relesed at \url{https://github.com/xjjxmu/TextRefiner}.
}
\end{abstract}

%

\section{Introduction}
{
Vision-Language Models (VLMs), such as CLIP~\cite{radford2021learning} and ALIGN~\cite{jia2021scaling}, have manifested exceptional generalization abilities, significantly benefiting a vast spectrum of applications like open-vocabulary image recognition~\cite{zhai2023sigmoid}, object detection~\cite{gu2022openvocabulary}, and image segmentation~\cite{li2022languagedriven}.
Generally, VLMs optimize an image encoder and a text encoder to align text and image features in a unified embedding space via contrastive loss.
By pre-training on large-scale web data, VLMs can acquire transferable representations in both visual and textual domains, facilitating their application to a broad range of downstream visual tasks. 
As such, numerous research efforts have been directed towards devising approaches to adapt VLMs in a manner mindful of both parameters and data~\cite{khattak2023maple,shen2024multitask,gao2024clip,yu2023task,li2024graphadapter}.

To date, the predominant focus of research has predominantly fallen on employing prompt learning~\cite{zhou2022learning,zhou2022conditional,kunananthaseelan2023lavip,khattak2023self}, a widely acknowledged strategy for adapting large pre-trained models.
As an initial foray, CoOp~\cite{zhou2022learning} leveraged prompt learning to adapt text prompts of CLIP, resulting in marked performance improvements in downstream tasks. 
Subsequent research efforts have sought to extend prompt learning into the visual~\cite{wu2023cora,li2024promptkd,shen2024multitask} and multimodal domains~\cite{khattak2023maple,xing2023dual,gao2024lamm,xin2024mmap}, aiming to thoroughly fine-tune the representations across modalities to enhance their alignment.

\begin{figure}
    \centering
    \includegraphics[width=\linewidth]{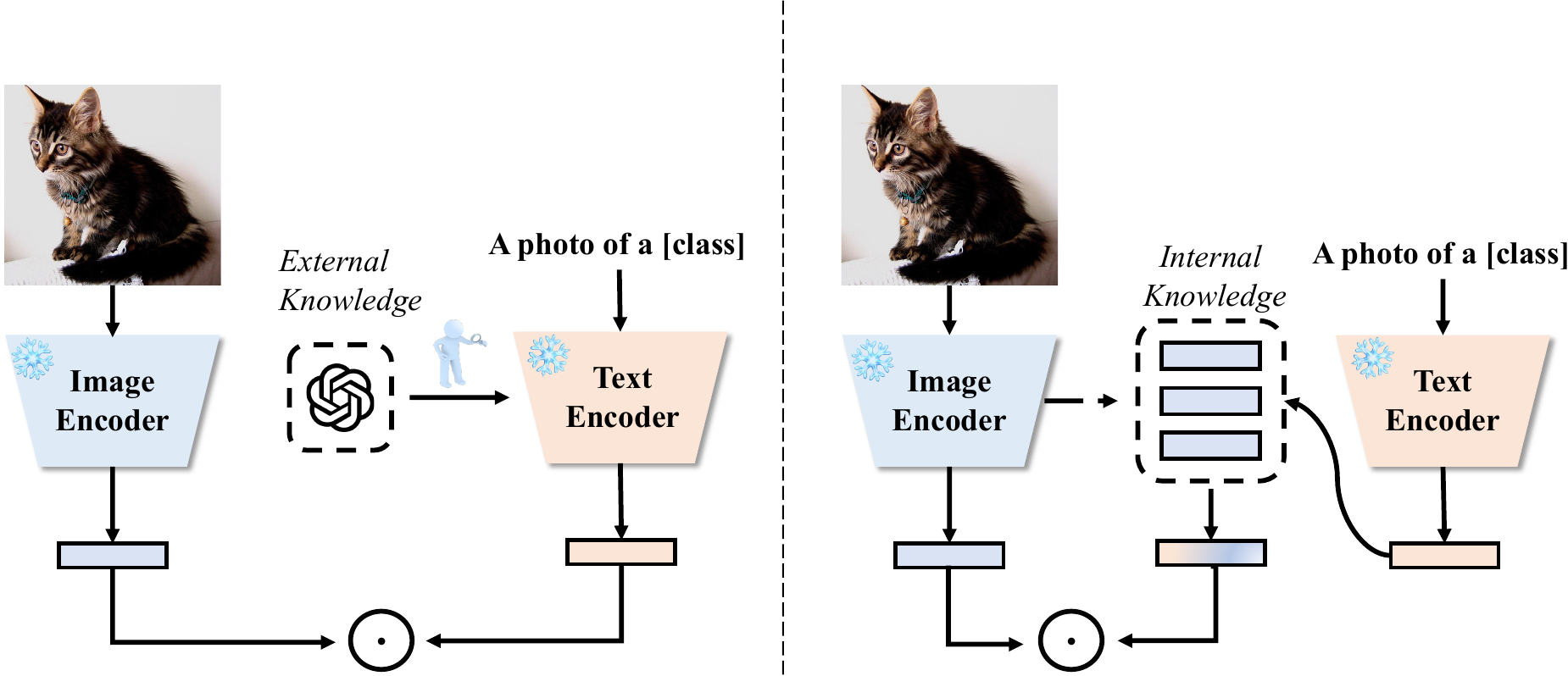}
    \caption{{Comparison of different paradigms for enriching text prompts. Previous methods (left) introduced \textbf{\textit{external}} knowledge experts to furnish fine-grained descriptions of each class, necessitating an extra filtering process to maintain alignment with downstream datasets. In contrast, our proposed TextRefiner (right) leverages the \textbf{\textit{internal}} knowledge of the image branch to supply fine-grained, localized region information, thereby drastically reducing the inference overhead while maintaining the performance.}}
    \label{fig:compare_llamp}
    \vspace{-1.5em}
\end{figure}

Despite the ongoing advancements, a prevalent limitation of these works is their tendency to enhance high-level semantics in a coarse-grained, global manner, resulting in holistic alignment across modalities. 
Consequently, the tailored prompts may fail to instruct the model to discern diverse visual concepts from local regions, thereby hindering its capability to generalize across classes that share similar visual attributes.
To overcome this limitation, recent works such as ArGue~\cite{Tian_2024_CVPR} and LLaMP~\cite{Zheng_2024_CVPR} utilize large language models (LLMs) to enrich the text prompts with fine-grained class descriptions that enhance transfer performance, as shown in Figure~\ref{fig:compare_llamp} (left).
Unfortunately, these approaches necessitate additional processes to sift through the LLMs outputs to ensure their relevance to downstream datasets, thereby introducing significant inference overhead.

This work aims to enrich VLM prompt tuning without the reliance on external knowledge experts, as previously described. 
Our core impetus stems from a widely accepted consensus in the traditional computer field: \textit{fine-grained visual concepts at local regions are prominently present in the intermediate layer outputs of the visual networks}~\cite{zeiler2014visualizing,kim22g,selvaraju2020grad}. As such, one can intuitively exploit the internal knowledge within the VLM, specifically the intermediate outputs of the image encoder, to deliver precise class descriptions and achieve local alignment without the need for external LLMs assistance.

To this end, we propose a plug-and-play method, dubbed TextRefiner, to refine the text prompt in VLM tuning without the need for external LLMs' assistance.
As depicted in Fig~\ref{fig:compare_llamp}, TextRefiner incorporates a novel local cache module to encapsulate fine-grained visual concepts derived from local tokens within the image branch. 
More particularly, amidst VLM tuning, the image branch continuously writes fine-grained semantic information into the cache in a class-wise manner. The fine-grained information retrieved for each category is concatenated with the corresponding class embedding, followed by a feature aggregation module to fuse global and local information to enhance the representation capability for the text branch. 
To further align the intermediate embeddings between modalities, we furnish TextRefiner with a feature alignment module to transform local tokens into the text embedding space.
Owing to such design, the fine-grained local features from the image branch can be seamlessly utilized to deliver class descriptions, thereby effectively refining the vanilla text prompting performance.

It is noteworthy that TextRefiner is both scalable and orthogonal to augment the efficacy of prevailing VLM tuning methods,~\emph{i.e}, the local cache module operates independently with existing prompt tuning paradigms.
For example, TextRefiner enhances CoOp by 5.30\% on 11 benchmarks, surpassing CoCoOp which introduced instance-specific features for text prompts to improve generalization. 
Equipped with TextRefiner, PromptKD achieves state-of-the-art performance, surpassing LLaMP, which relies on external experts, by 1.06\%. 
Moreover, TextRefiner demonstrates high inference efficiency. CoCoOp achieves 20.45 FPS, LLaMP reaches 1473.46 FPS, while PromptKD \emph{w/}TextRefiner achieves 12793.26 FPS. This efficiency is due to its reliance on simple matrix multiplication at the text output, unlike CoCoOp and LLaMP, which require additional modules to extract and combine information with the text input. 

The contributions of this paper are delineated as follows:
\begin{itemize}
    \item We introduce TextRefiner, which, for the first time, exploits the internal knowledge of the image branch to refine the text prompts of VLMs.
    \item TextRefiner introduces three principal innovations, encompassing local cache, feature aggregation, and feature alignment, which collaboratively enhance the incorporation of fine-grained visual attributes into the text prompts.
    \item Extensive experiments demonstrate the effectiveness of TextRefiner in boosting VLM prompt tuning, even without additional inference costs associated with prior methods that utilize external knowledge from LLMs.
\end{itemize}

}

\section{Related Works}
\subsection{Vision-Language Model}
{
Traditional visual representation learning relied exclusively on a predefined set of labels limiting its expansion to unseen classes~\cite{deng2009imagenet,he2016deep,dosovitskiy2021an,liu2022convnet}. 
Vision-language models redress this deficiency by incorporating natural language supervision, thereby aiding vision systems in mastering more visual concepts. These models are pre-trained on copious amounts of raw text paired with images from the Internet in a self-supervised manner.
Marked advances such as ALIGN~\cite{jia2021scaling} and CLIP~\cite{radford2021learning} utilize contrastive loss during pre-training to align both modalities into a unifying embedding space.
Benefiting from the large-scale dataset and contrastive training objective, the learned visual representations connect with language and are highly generalizable, which has been widely applied to various downstream vision tasks, such as object detection~\cite{gu2022openvocabulary}, semantic segmentation~\cite{li2022languagedriven} and depth estimation~\cite{zhang2022can}.
Despite these formidable capabilities, the transfer of VLMs to downstream tasks at few-shot scenarios poses a formidable challenge, where the quantity of data in such tasks falls substantially short relative to pre-training, leading to a significant falloff in generalizability~\cite{zhu2023prompt,khattak2023self,CoPrompt}. Our proposed TextRefiner in this paper is crafted to achieve efficient transfer learning in low-data scenarios while preserving generalization.
}
\subsection{Prompt Learning in VLMs}
{Prompt learning is progressively favored in adapting Vision-and-Language Models (VLMs) to downstream tasks without necessitating a full re-training of the original model.
For VLMs, prompt learning entails introducing learnable text or visual prompts that transcend manually-defined prompts like \texttt{a photo of a 
\{class\}}.
CoOp~\cite{zhou2022learning} incorporates learnable tokens into predefined text prompts within the text branch. These tokens, universally shared across classes, glean task-related knowledge to align with the image features derived from the downstream dataset.
Zhou~\emph{et al.}~\cite{zhou2022conditional} proposed conditional prompt learning (CoCoOp), constraining image features in an instance-specific manner to curtail overfitting in few-shot scenarios and bolster generalization towards unseen classes.
PromptSRC~\cite{khattak2023self} further modulates learned prompts utilizing pre-trained features to enhance generalization.
MaPLe~\cite{khattak2023maple} expands prompt learning into multimodal branches, enabling the prompt to not only capture characteristics within the text branch but also enhance inter-modal alignment.
Despite their effectiveness, these methods generally tune the prompt on a rather coarse granularity scale, which tends to overlook the fine-grained image feature, potentially impairing the generalization across classes that share similar visual attributes.
To mitigate this, ArGue~\cite{Tian_2024_CVPR} and LLaMP~\cite{Zheng_2024_CVPR} utilize large language models (LLMs) to provide detailed attribute information of classes to the text branch, achieving fine-grained alignment in downstream tasks.
However, such LLM-based methods necessitate well-curated filtering of knowledge generated by the LLMs to ensure its relevance to downstream tasks, consequently incurring additional computational expenses.
Our method capitalizes on the semantic information embedded in image local tokens to realize fine-grained alignment without the need for complex filtering or huge inference costs.
}
\section{Method}
\subsection{Background} 
{\textbf{Preliminaries.}}
{By aligning image and text in a joint embedding space, Vision-Language Models (VLMs) like CLIP~\cite{radford2021learning} and ALIGN~\cite{jia2021scaling} mark significant advancements in vision applications.
Following previous works~\cite{zhou2022conditional,khattak2023self,Zheng_2024_CVPR}, this paper utilizes CLIP as the foundation model, with relevant preliminary details provided herein.
CLIP consists of an image encoder, labeled as $f_I$, and a text encoder referred to as $f_T$. Both encoders incorporate a feature extractor and a projection layer the transfer the multimodal inputs into a joint embedding space. Without loss of generality, we utilize the CLIP model equipped with a vision transformer (ViT) for illustration.

During the training phase, a contrastive loss~\cite{chen2020simple} is employed to maximize the cosine similarity between embeddings of corresponding image-text pairs $\{I,T\}$, where $I$ denotes the image and $T$ the associated caption. For testing, CLIP integrates each class name within a hard prompt formatted as \texttt{a photo of a \{class\}}, creating textual class descriptions $\textbf{P} \in \mathbb{R}^{C \times l}$, where $C$ is the number of classes and $l$ is the harmonized length of text sequences. 
The text encoder $f_T$ then generates textual embeddings $\textbf{E} \in \mathbb{R}^{C \times d}$ for $\textbf{P}$, which act as class templates.
Then, with the image feature $v \in \mathbb{R}^{d}$ outputted by the image encoder $f_I$, the prediction probabilities are determined by the corresponding similarity between textual embeddings and the image feature, expressed as:
\begin{equation}
    P(c) = 
    \frac{\text{exp}(\text{cos}(v,\textbf{E}_c)/\tau)}
    {\sum_{c=1}^{C}{\text{exp}(\text{cos}(v,\textbf{E}_c)/\tau)}},
\end{equation}
where $c$ represents a specific class, cos(\,$\cdot$\,) calculates the cosine distance between vectors, and $\tau$ is a tuning parameter for scaling the softmax function.

\textbf{VLMs Prompt Tuning.} Instead of manually crafted hard prompts, VLM prompt tuning methods, pioneered by CoOp~\cite{zhou2022learning}, learn soft prompts to further improve the transferring performance on downstream tasks.
Specifically, CoOp incorporates multiple learnable word embeddings \(p_i\) shared within all classes to construct textual descriptions as \(\textbf{P} =  [p_1,p_2,...,p_L]\texttt{\{class\}} \), where $L$ specifies the number of soft prompts. With these learnable prompts, CLIP can enhance task-relevant textual descriptions across all classes. Subsequent research efforts introduce learnable context tokens into the image branch or both branches to better learn downstream representations~\cite{wu2023cora,shen2024multitask,khattak2023maple,xing2023dual}. Despite their efficacy, the prevailing paradigms largely enhance embeddings in a coarse-grained manner, with prompts that are universal across all classes, achieving only global alignment and often failing to reflect the diverse visual attributes of localized regions.
Although recent works have revealed the potential in utilizing external knowledge from LLMs~\cite{touvron2023llama,touvron2023llama2} to refine the learned prompts with fine-grained class-wise information, it comes at additional inference costs associated with filtering the outputs of LLMs to ensure their pertinence to specific downstream datasets.

\begin{figure*}[!htbp]
    \centering
    \includegraphics[width=\linewidth]{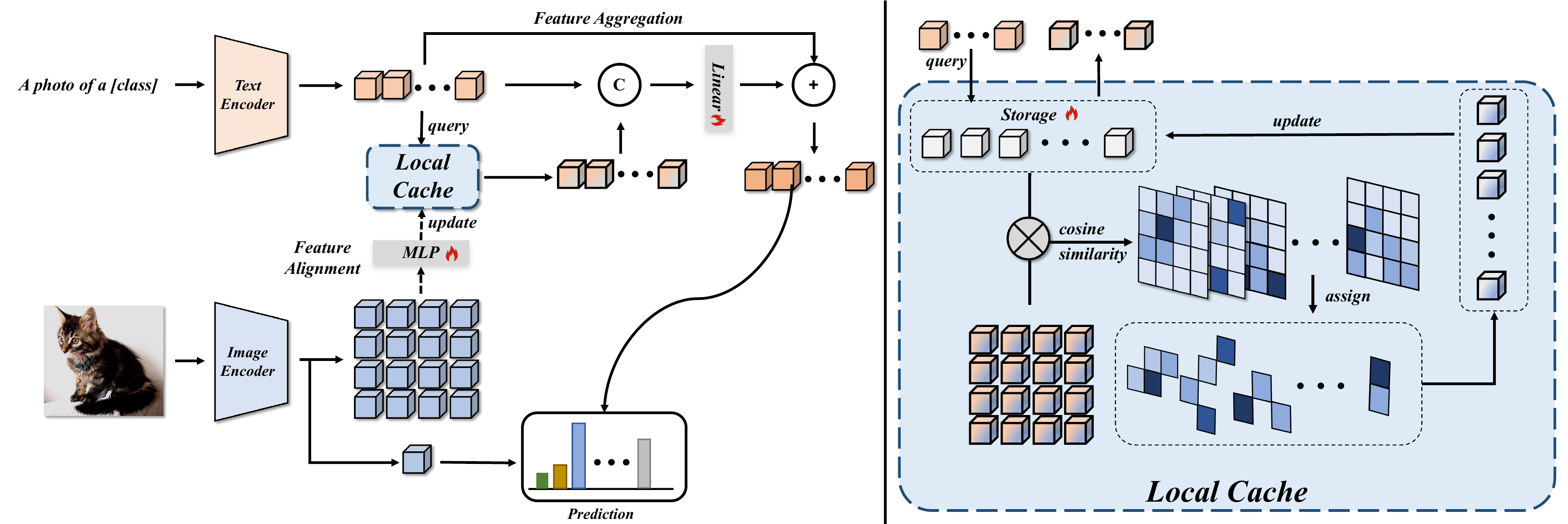}
    \caption{The framework of TextRefiner, which is composed of local cache, feature aggregation and feature alignment. Here, each item in the local cache can be considered as an attribute prior which will be updated by local tokens from the image branch. Therefore, textual class embedding can obtain corresponding linguistic visual attributes by querying this cache.}
    \label{fig:framework}
\end{figure*}

\subsection{TextRefiner}

To mitigate the aforementioned challenges, this paper presents a plug-and-play method to refine the text prompt in VLM prompt tuning without depending on external assistance, termed TextRefiner. 
The core contribution of TextRefiner lies in enhancing the textual embedding with linguistic visual attributes using local features within the image branch, which have been demonstrated to encompass rich fine-grained visual concepts at local regions~\cite{ghiasi2022vision,ma2023visualizing,kim2022vit}.
TextRefiner achieves this objective through three complementary innovations, including local cache, feature aggregation and feature alignment, as shown in Fig.~\ref{fig:framework}. We detailedly introduce them as below.

\subsubsection{Local Cache.}
The visual encoder ViT splits input image into non-overlapping patches and feeds them into stacked transformer blocks to obtain a series of local tokens \(\textbf{V} = \{ v_1, v_2, ... ,v_N\}\), where \( N \) represents the number of patches. 
These local tokens have been widely demonstrated to be capable of capturing specialized visual features such as edges, textures, and represent discernible conceptual categories~\cite{ghiasi2022vision}.
Therefore, we propose to leverage these internal local tokens within VLMs as visual attribute descriptions for classes, forsaking the practice of using external experts as previous methods do.
To achieve this, we propose a novel local cache module to store the fine-grained visual concepts from local tokens. 
Instead of coarsely storing all local tokens in bulk, we establish a fixed number of entries within the local cache and cluster similar local tokens into a single entry to selectively capture corresponding visual attributes. For instance, when processing a zebra sample, our objective is to amalgamate its pronounced black and white stripe features into a cache entry that epitomizes textures. To this end, we construct an storage \( \mathbf{A} \in \mathbb{R}^{M \times d}\), where $M$ entries in \( \mathbf{A} \) 
are used to collect information from local tokens across all instances based on similarity and \( d \) denotes the feature dimension. 
Given local tokens, we calculate the cosine similarity between the attribute storage and local tokens, followed by a softmax function to obtain probability, as follows:
\begin{equation}
    \textbf{D}_{i,j} = \frac{\text{exp}(\text{cos}(v_i, \textbf{A}_j))}{\sum_{j=1}^{M}{\text{exp}(\text{cos}(v_i, \textbf{A}_j))}}.
\end{equation}
Then, the local tokens are assigned to the entry in \( \textbf{A}\) with the highest probability. For example, the allocated position of the \( j\)-th entry in the local cache is formulated as follows:
\begin{equation}
    G_j = \{\,i\,| \mathop{\arg\max}\limits_{k} \textbf{D}_{i,k} = j\}.
\end{equation}
Subsequently, the \( j\)-th entry gathers information from local tokens as:
\begin{equation}
    \textbf{A}_j = \gamma \cdot \textbf{A}_j + (1-\gamma) \sum_{i \in G_j} \textbf{D}_{i,j} \cdot v_i ,
\end{equation}
where \( \gamma \) denotes the momentum coefficient. Through the above updates, we achieve continuously memorizing the fine-grained information inherent in the local tokens and this information subsequently can be utilized to enhance the descriptive capabilities of text representation. 

\subsubsection{Feature Aggregation.}
To harness the potential of the fine-grained information in the local cache, we introduce feature aggregation operation. 
As shown in Figure~\ref{fig:framework}, we enhance text embedding of classes by integrating local context from the local cache into the global context from CLIP via residual connection~\cite{he2016deep,yu2023task}.
Specifically, we begin by matching 
each text embedding \(\textbf{E}_i\) from CLIP with the corresponding visual attributes stored in entries of the local cache as:
\begin{equation}
    \textbf{W}_{i,j} = \frac{\text{exp}(\text{cos}(\textbf{E}_i, \textbf{A}_j))}{\sum_{j=1}^{M}{\text{exp}(\text{cos}(\textbf{E}_i, \textbf{A}_j))}},
\end{equation}
Then, the detailed descriptive embedding associated with the corresponding \( i\)-th class is calculated as:
\begin{equation}
    \overline{\textbf{E}}_i = \sum_{j=1}^{M} \textbf{W}_{i,j} \cdot \textbf{A}_j.
\end{equation}
Then, we concatenate the detailed descriptive embedding with the original text embedding and subsequently employ a linear layer with residual connection to aggregate the two types of features:
\begin{equation}
    \label{eq:agg}
    \hat{\textbf{E}}_i = \alpha \cdot \text{Linear}([\textbf{E}_i, \overline{\textbf{E}}_i]) + \textbf{E}_i ,
\end{equation}
where \(\alpha\) is the coefficient hyperparameter.

\subsubsection{Feature Alignment.}
CLIP aligns the global feature from images and texts to learn a joint embedding space. However, local features in images lack explicit alignment, leading to the modality gap between the local feature \(\textbf{V}\) and the textual embedding \(\textbf{E}\). To address potential feature shift, we further incorporate a feature alignment module to transform local features into the text embedding space, mitigating the modality gap. The feature alignment module simply consists of a 2-layer MLP where \(\textbf{V}\) will be transformed as:
\begin{equation}
    \hat{\textbf{V}} = {\textbf{W}_2}\sigma(\text{norm}(\textbf{W}_1\textbf{V})).
\end{equation}

\begin{table*}[htb!]
    \small
    \centering

    \begin{subtable}[t]{\textwidth}
        \begin{tabular}{cccc|ccc|ccc|ccc}
    \toprule \multirow{2}[3]{*}{ Method } & \multicolumn{3}{c}{ \textit{Average} } & \multicolumn{3}{c}{ ImageNet} & \multicolumn{3}{c}{ Caltech101} & \multicolumn{3}{c}{ OxfordPets} \\
    \cmidrule(lr){2-4}\cmidrule(lr){5-7}\cmidrule(lr){8-10}\cmidrule(lr){11-13} 
     & Base & Novel & HM & Base & Novel & HM & Base & Novel & HM & Base & Novel & HM \\
     \midrule
        CLIP  & 69.34 & 74.22 & 71.70 & 72.43 & 68.14 & 70.22 & 96.84 & {94.00} & 95.40 & 91.17 & 97.26 & 94.12\\
        CoOp & 82.69 & 63.22 & 71.66 & {76.47} & 67.88 & 71.92  & 98.00 & 89.81 & 93.73 & 93.67 & 95.29 & 94.47  \\
        CoCoOp  & 80.47 & 71.69 & 75.83 & 75.98 & {70.43} & {73.10} & 97.96 & 93.81 & {95.84} & 95.20 & 97.69 & {96.43}\\
      PromptSRC & 84.26 & 76.10 &79.97 &77.60 &70.73 &74.01&98.10 &94.03 &96.02 &95.33 &97.30 & 96.30 \\
      MaPLe & 82.28 & 75.14 & 78.55 &76.66 &70.54 &73.47 &97.74 &94.36 & 96.02 &95.43 &97.76 &96.58 \\ 		
      PromptKD  & 84.11 & 78.28 & 81.09 &77.63	&70.96	& 74.15 & 98.31 &	96.29 &	97.29 &	93.42 &	97.44 &	95.39 \\
     LLaMP & 85.16  & 77.71 & 81.27 & \textbf{77.99} & 71.27 & \textbf{74.48} & 98.45 & 95.85 & 97.13 & \textbf{96.31} & 97.74  & \textbf{97.02}  \\
    \midrule		
    \rowcolor{mygray} 
     CoOp \emph{w}/TextRefiner & 79.74 & 74.32 & 76.94 & 76.84&	70.54 & 73.56 &	98.13 & 94.43 & 96.24 &	95.27 &	97.65 &	96.45 \\	
     \rowcolor{mygray} 
     PromptKD \emph{w}/TextRefiner & \textbf{85.22} & \textbf{79.64} & \textbf{82.33} &  77.51 &	\textbf{71.43} & 74.35 & \textbf{98.52} & \textbf{96.52} & \textbf{97.51} & 95.60 & \textbf{97.90} &	96.74 \\
    \bottomrule
    \end{tabular}
    \end{subtable}

    \begin{subtable}[t]{\textwidth}
        \centering
        \vspace{5pt}
        \begin{tabular}{cccc|ccc|ccc|ccc}
    \toprule \multirow{2}[3]{*}{ Method } & \multicolumn{3}{c}{StanfordCars} & \multicolumn{3}{c}{ Flowers102} & \multicolumn{3}{c}{ Food101} & \multicolumn{3}{c}{ FGVCAircraft} \\
    \cmidrule(lr){2-4}\cmidrule(lr){5-7}\cmidrule(lr){8-10}\cmidrule(lr){11-13} 
     & Base & Novel & HM & Base & Novel & HM & Base & Novel & HM & Base & Novel & HM \\
     \midrule
     CLIP  & 63.37 & 74.89 & 68.65 & 72.08 & 77.80 & 74.83 & 90.10 & 91.22 & 90.66 & 27.19 & 36.29 & 31.09  \\
    CoOp & 78.12 & 60.40 & 68.13 & 97.60 & 59.67 & 74.06 & 88.33 & 82.26 & 85.19 & 40.44 & 22.30 & 28.75 \\
    CoCoOp & 70.49 & 73.59 & 72.01 & 94.87 & 71.75 & 81.71 & 90.70 & 91.29 & 90.99 & 33.41 & 23.71 & 27.74 \\
    PromptSRC & 78.27 & 74.97 &76.58 &98.07 &76.50 &85.95&90.67 &91.53 &91.10 &42.73 &37.87 & 40.15 \\
    MaPLe & 72.94 & 74.00 & 73.47 &95.92 &72.46 &82.56 &90.71 &92.05 & 91.38 &37.44 &35.61 &36.50 \\		
    PromptKD  &80.48 & 81.78 & 81.12 & 98.69 & 81.91 &	89.52 & 89.43 & 91.27 & 90.34 &	43.61 & 39.68 & 41.55 \\
    LLaMP & \textbf{81.56} & 74.54 & 77.89 & 97.82 & 77.40 & 86.42 & 91.05 & 91.93 & 91.49 & \textbf{47.30} & 37.61 & 41.90 \\
    \midrule 		
    \rowcolor{mygray} 
     CoOp \emph{w}/TextRefiner & 71.40 & 70.90 & 71.15 & 95.92 & 74.33 & 83.76 & 90.88 & 91.43 & 91.15 &	35.35 & 35.87 & 35.61 \\
     \rowcolor{mygray} 
     PromptKD \emph{w}/TextRefiner & 80.91 & \textbf{81.83} & \textbf{81.37}& \textbf{99.30} & \textbf{82.91} & \textbf{90.37} & \textbf{91.42} &\textbf{92.71} &	\textbf{92.06} &	45.01	& \textbf{40.12}	& \textbf{42.42} \\
    \bottomrule
    \end{tabular}
    \end{subtable}

    \begin{subtable}[t]{\textwidth}
        \centering
        \vspace{5pt}
        \begin{tabular}{cccc|ccc|ccc|ccc}
    \toprule \multirow{2}[3]{*}{ Method } & \multicolumn{3}{c}{ SUN397 } & \multicolumn{3}{c}{  DTD} & \multicolumn{3}{c}{  EuroSAT} & \multicolumn{3}{c}{ UCF101} \\
    \cmidrule(lr){2-4}\cmidrule(lr){5-7}\cmidrule(lr){8-10}\cmidrule(lr){11-13} 
     & Base & Novel & HM & Base & Novel & HM & Base & Novel & HM & Base & Novel & HM \\
     \midrule
        CLIP  & 69.36 & 75.35 & 72.23 & 53.24 & 59.90 & 56.37 & 56.48 & 64.05 & 60.03 & 70.53 & 77.50 & 73.85 \\
        CoOp  & 80.60 & 65.89 & 72.51 & 79.44 & 41.18 & 54.24 & 92.19 & 54.74 & 68.69 & 84.69 & 56.05 & 67.46 \\
        CoCoOp & 79.74 & 76.86 & 78.27 & 77.01 & 56.00 & 64.85 & 87.49 & 60.04 & 71.21 & 82.33 & 73.45 & 77.64 \\
        PromptSRC & 82.67 & 78.47 &80.52 &83.37 &62.97 &71.75 & 92.90 &73.90&82.32 &87.10 &78.80 & 82.74 \\
        MaPLe & 80.82 & 78.70 & 79.75 &80.36 &59.18&68.16 &\textbf{94.07} &73.23 & 82.35 &83.00 &78.66 &80.77 \\
      PromptKD  & 82.53 & \textbf{80.88} & 81.70 & 82.86 & 69.15 & 75.39 & 92.04 & 71.59 &	80.54 & 86.23 & 80.11 & 83.06 \\
     LLaMP &\textbf{83.41}  & 79.90 & 81.62 & 83.49 & 64.49 & 72.77 &91.93  & \textbf{83.66} & \textbf{87.60} & 87.13 & 80.66 & 83.77 \\
    \midrule
    \rowcolor{mygray} 
     CoOp \emph{w}/TextRefiner &80.96	& 76.49	 &78.66 & 75.35 & 58.09 & 65.60 &	74.57 & 72.82 & 73.68 &	82.52 & 75.01 & 78.59 \\
     \rowcolor{mygray} 
     PromptKD \emph{w}/TextRefiner & 83.02 & 80.50 & \textbf{81.74} & \textbf{83.91} & \textbf{71.01} & \textbf{76.92} & 92.99 & 79.22 & 85.55& \textbf{89.20} & \textbf{81.90} & \textbf{85.39} \\		
    \bottomrule
    \end{tabular}
    \end{subtable}

    \caption{Comparison between our method and other existing methods on base-to-novel generalization.}
    \vspace{-1.5em}
    \label{tab:b2n}
\end{table*}

\subsection{Training}
During the training, we adhere to the original CLIP framework, utilizing contrastive loss as the primary classification supervision, formulated as follows:
\begin{equation}
    \mathcal{L}_{cls} = -\log{
    \frac{\exp(\cos(v,\hat{\textbf{E}}_c)/\tau)}
    {\sum_{j=1}^{C}{\exp(\cos(v,\hat{\textbf{E}}_j)/\tau)}}
    }.
\end{equation}
In addition to the vanilla contrastive loss, we augment TextRefiner with a semantic loss and a regularization loss component. 
Respectively, we select the top-k transformed local features with attention scores and calculate the average of all tokens to obtain \(\mathbf{S} \in \mathbb{R}^{(k+1) \times d}\). The semantic loss ensures that $\textbf{S}$ align with their corresponding text embeddings, expressed as:
\begin{equation}\label{eq:sem_loss}
    \mathcal{L}_{sem} = \frac{1}{k+1}\sum_{i=1}^{k+1}{\log{
        \frac{\exp(\cos( {\textbf{S}}_i,\hat{\textbf{E}}_c)/\tau)}
        {\sum_{j=1}^{C}{\exp(\cos({\textbf{S}}_i,\hat{\textbf{E}}_j)/\tau)}}
        }}.
\end{equation}
On the other hand, we implement a regularization loss to curb overfitting to the limited amount of training images in VLMs fine-tuning scenarios, which can be denoted as:
\begin{equation}
    \mathcal{L}_{reg} = |\textbf{E} - \hat{\textbf{E}}|.
\end{equation}
The overall loss can be formulated as follows:
\begin{equation}\label{eq:final_loss}
    \mathcal{L} = \mathcal{L}_{cls} + \lambda_1 * \mathcal{L}_{sem} + \lambda_2 * \mathcal{L}_{reg},
\end{equation}
where $ \lambda_1 $ and $ \lambda_2 $ are hyperparameters utilized to balance the loss components.

\subsection{Inference}
During the inference phase for the input image denoted by $x$, we compute the cosine similarity between \(v=f_I(x)\) and the fused textual embedding \(\hat{\textbf{E}}_i\) for all classes. The predicted label $y$ is ascertained based on the maximum likelihood, expressed as:
\begin{equation}
    y = \mathop{\arg\max}\limits_{i}{\frac{\text{exp}(\text{cos}(v,\hat{\textbf{E}}_i)/\tau)}
    {\sum_{j=1}^{C}{\text{exp}(\text{cos}(v,\hat{\textbf{E}}_j)/\tau)}}}
\end{equation}
Unlike other methods that insert hard or carefully designed learnable prompts representing visual attributes into the input embedding of the text branch, our approach only requires feature aggregation between the textual output embeddings and the local cache, providing an efficiency advantage.
}

\section{Experiment}

\begin{table}[htbp!]
    \centering
    \scalebox{0.9}
    {
    \begin{tabular}{lccccc}
    \toprule 
    \multirow{2}{*}{ Method } & Source &\multicolumn{4}{c}{Target} \\
    \cmidrule(lr){2-2}\cmidrule(lr){3-6}
    & ImageNet & -V2 & -Sketch & -A & -R \\
    \midrule
    CLIP& 66.73 &60.83 &46.15 &47.77 &73.96 \\
    CoOpOp&  71.02 & 64.07 &48.75 &50.63 & 76.18 \\
    PromptSRC& 71.27 & 64.35 & 49.55 & 50.90&  
    77.80  \\
    \midrule
    CoOp& 71.51 &64.20& 47.99 &49.71 &75.21  \\
    \rowcolor{mygray}
    \quad+ TextRefiner& 72.06 & 65.02 & 48.58 & 49.77 & 76.30  \\
    
    MaPLe & 70.72& 64.07 & 49.15 & 50.90 & 76.98  \\
    \rowcolor{mygray}
    \quad+ TextRefiner& 71.13 & 64.54 & 49.08& 51.49& 77.71 \\
    \bottomrule
    \end{tabular}
    }
    \caption{Comparison between our method and other existing methods on cross-domain generalization. Models will be trained with 16-shot ImageNet and test in out-of-distribution datasets.}
    \label{tab:xd}
    \vspace{-1.5em}
\end{table}
\subsection{Settings}
\subsubsection{Datasets.}  
{We follow the common practice in the literature~\cite{zhou2022conditional,khattak2023self,khattak2023maple} to adopt two experimental settings: \textit{base-to-novel} and \textit{cross-domain}. 
For \textit{base-to-novel} evaluation, we adopt a wide range of visual recognition benchmark datasets \emph{i.e.}, ImageNet~\cite{deng2009imagenet}, Caltech101~\cite{fei2004learning}, OxfordPets~\cite{parkhi2012cats}, StanfordCars~\cite{krause20133d}, Flowers102~\cite{nilsback2008automated}, Food101~\cite{bossard2014food}, FGVCAircraft~\cite{maji2013fine}, SUN397~\cite{xiao2016sun}, DTD~\cite{cimpoi2014describing}, EuroSAT ~\cite{helber2019eurosat}and UCF101~\cite{soomro2012ucf101}. In this setting, the datasets are split into two non-overlapping sets where the base classes and novel classes serve as training sets and test sets, respectively.
For cross-domain evaluation, we train our model on ImageNet and test the generalization performance on its variants: ImageNetV2~\cite{recht2019imagenet}, ImageNet-Sketch~\cite{wang2019learning}, ImageNet-A~\cite{hendrycks2021natural} and ImageNet-R~\cite{hendrycks2021many}.
Both are trained using 16 shots per class from the training sets and are tested on the full test sets.
\subsubsection{Implementation details.}
For a fair comparison, we adopt CLIP with ViT-B/16 vision encoder and verify the effectiveness of our method on CoOp~\cite{zhou2022learning}, MaPLe and PromptKD~\cite{li2024promptkd}. 
The setting of epochs, batch size, learning rate, optimizer and soft token length corresponds with the original papers, except for CoOp, where the number of epochs was reduced from 200 to 10. 
In particular, the momentum $\alpha$ and the fusion factor $\beta$ are configured to be 0.8 and 0.2, respectively. For the hyper-parameters in Eq.~\ref{eq:final_loss}, we set $\lambda_1$ to 0.02 and $\lambda_2$ to 20 based on empirical observation.
}
For simplicity, we report ablation results on CoOp to verify its effectiveness.
\subsection{Main Results}
\subsubsection{Base-to-novel.}
Firstly, we report the quantitative results of various methods for   \textit{base-to-novel} tasks on 11 benchmarks. Table~\ref{tab:b2n} shows our method can consistently enhance the performance of existing methods. In particular, 
TextRefiner enhances CoOp's generalization on novel classes, improving accuracy from 63.22\% to 74.32\%. Compared to CoCoOp, which integrates instance-conditional context produced by meta-net to refine text prompt, TextRefiner effectively local tokens and  offers additional improvements.
For PromptKD, our method enhances their performance on fine-grained benchmarks. On \emph{DTD}, TextRefiner increases novel accuracy from 69.15\% to 71.01\%, On \emph{EuroSAT}, TextRefiner achieves improvements of 7.63\%.
Compared to LLaMP, which uses LLMs to provide detailed descriptions according to class names, our method achieves better generalization and provides a better balance between base and novel classes, where PromptKD \emph{w/}TextRefiner achieves average gains of 1.97\% in novel accuracy, and 1.06\% in harmonic mean on average. This suggests that additional filtering modules run the risk of overfitting downstream tasks, requiring careful design. In contrast, our approach leverages the internal knowledge in visual networks to improve downstream performance while enhancing generalization.
These improvements verify that our method can achieve better balances on seen and unseen data, thus showcasing stronger generalization compared to the original methods.

\subsubsection{Cross-domain.}
Moving beyond fundamental \textit{base-to-novel} benchmarks, we exploit the generalization capacity of TextRefiner across domains on four commonly-used datasets \textit{i.e.}, ImageNet-V2, ImageNet-Sketch, ImageNet-A and ImageNet-R.
The experimental results are shown in Table~\ref{tab:xd}, where a significant improvement has been observed in all scenarios and the robustness to domain shift is verified.

\subsubsection{Efficiency.}
\begin{figure}
    \centering
    \includegraphics[width=\linewidth]{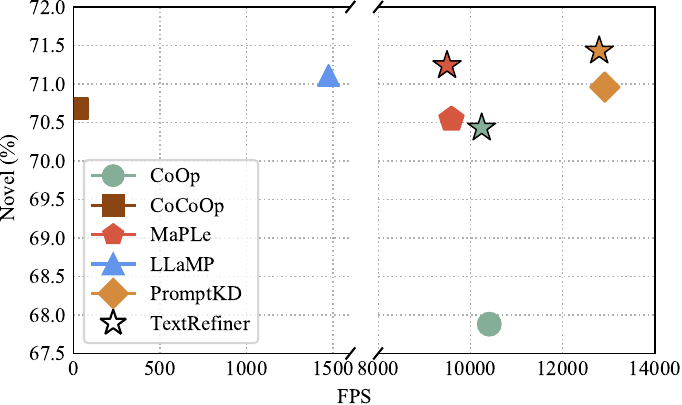}
    \caption{Comparison of inference efficiency among existing methods on the ImageNet dataset. Our TextRefiner is more efficient than LLaMP which relies on external knowledge experts to furnish fine-grained descriptions of each class.}    
    \label{fig:fps}
    \vspace{-1em}
\end{figure}
We also provide a comparison of inference efficiency which is evaluated with one single A800 GPU based on the officially released code. As shown in Figure~\ref{fig:fps}, TextRefiner only slightly reduces FPS and remains highly efficient. In contrast, LLaMP, which relies on external experts, significantly compromises inference speed.

\subsection{Ablations}
\subsubsection{The effects of each component.}
\begin{table}[]
    \centering
\begin{tabular}{cccccc}
\toprule 
 \multirow{2}{*}{TextRefiner} & \multirow{2}{*}{ $\mathcal{L}_{sem}$ } & \multirow{2}{*}{ $\mathcal{L}_{reg}$ } & \multicolumn{3}{c}{ ImageNet } \\
 \cmidrule{4-6}
 & & & Base & Novel & HM  \\
\midrule
              &   &    & 76.47 & 67.88 & 71.92 \\
  \checkmark  &   &    & 76.50 & 70.48 & 73.37  \\
  \checkmark  &  \checkmark  &    & 76.58 & 70.62 & 73.48  \\
  \checkmark  &    &  \checkmark  & 76.70 & 69.67 & 73.02  \\
  \rowcolor{mygray}
  \checkmark  &  \checkmark  &  \checkmark  & 76.84 & 70.54 & 73.56  \\

\bottomrule

\end{tabular}
    \caption{Ablation study on each component.}
    \label{tab:each}
\end{table}

To verify the effectiveness of the proposed method, we examine different combinations of components. As shown in Table \ref{tab:each}, TextRefiner can significantly enhance the performance of CoOp by providing fine-grained information in the local cache. 
The introduced semantic loss ensures that the semantic information stored in the local cache is distinguish and aligned with the text, further improving performance. Introducing regularization loss enhances performance on seen classes but restricts the utilization of fine-grained information in the local cache, which leads to decreased generalization on unseen classes. Utilizing all components together effectively balances the recognition of seen classes and the generalization of unseen classes.

\subsubsection{The effects of $M$.} 
\begin{figure}[!t]
    \centering
    \includegraphics[width=\linewidth]{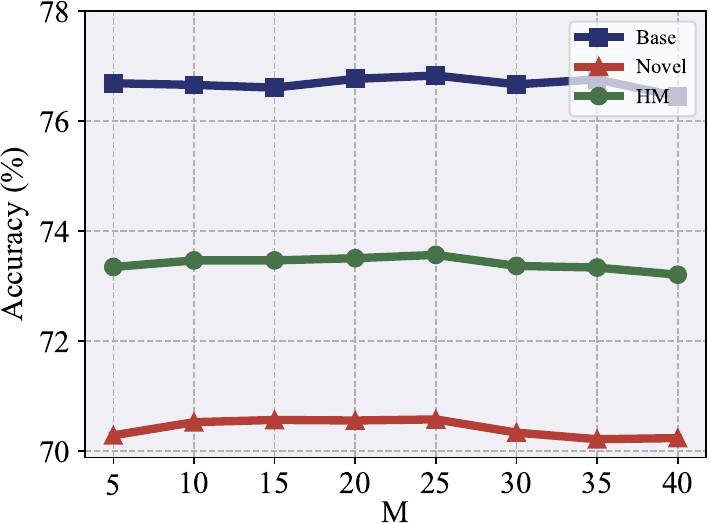}
    \caption{Ablation study on $M$ in $\textbf{A}$.}
    \label{fig:mem}
\end{figure}
We investigate the effect of $M$ in $\textbf{A}$, controlling the condensation level of local tokens, 
where a larger M means more entries to summarize the visual attributes from local tokens. As shown in Figure~\ref{fig:mem}, recognition performance improves as $M$ increases, but once $M$ reaches a certain value, performance starts to decline. This can be understood as follows: when $M$ exceeds a certain threshold, complete visual attributes are split and stored in multiple entries compromising their integrity and text embedding struggles to recombine the correct attributes.

\subsubsection{The effects of \(\lambda_1\) and \(\lambda_2\).}
\begin{table}[]
    \centering
    \begin{tabular}{cccccc}
\toprule
\(\frac{1}{\lambda_1}\) & 20 & 30 & 40 & 50 & 60  \\
\midrule
\text { Base } & 76.33 & 76.45 & 76.52 & 76.84 & 76.60  \\
\text { Novel } & 70.14 & 70.04 & 70.28 & 70.54 & 70.27  \\
\text { HM } &  73.10& 73.10 & 73.27 & 73.56 & 73.30    \\
\midrule 
\(\lambda_2\) & 5 & 10 & 15 & 20 & 25 \\
\midrule 
\text { Base } & 76.65 & 76.70 & 76.67 & 76.84 & 76.65  \\
\text { Novel } & 70.49 & 70.51 & 70.32 & 70.54 & 70.34  \\
\text { HM } & 73.44 & 73.47 & 73.36 & 73.56& 73.36   \\

\bottomrule
    \end{tabular}
    \caption{Ablation study on loss factors.}
    \label{tab:loss}
    \vspace{-1.5em}
\end{table}
To better understand the effects of these two loss factors, we fixed one and observed the performance changes as the other varied. As shown in Table 4,
the performance of our method consistently increases first and then decreases when the loss balance factor \(\lambda_1\) and \(\lambda_2\) increase. 
This observation suggests that excessively aligning local tokens can cause the stored information to overfit the local features of seen samples, leading to a spurious relationship. In addition, applying a strong regularization results in insufficient learning for TextRefiner.

\subsubsection{The effects of fusion coefficient \(\alpha\).}
\begin{figure}[!t]
    \centering
    \includegraphics[width=\linewidth]{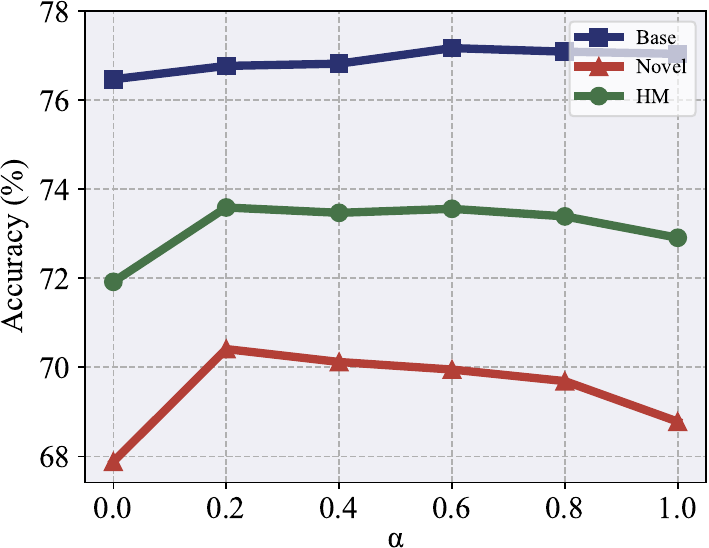}
    \caption{Ablation study on aggregation coefficient in Eq.~\ref{eq:agg}.}
    \label{fig:alpha}
\end{figure}
We also studied the hyperparameter \(\alpha\), where
a larger \(\alpha\) indicates the final text embedding contains more local information from $\overline{\textbf{E}}$. 
In Figure~\ref{fig:alpha}, as \(\alpha\) increases, accuracy on the base class continuously improves, while accuracy on the novel class would increase first and then continuously decrease. 
This indicates that with a limited number of samples, over-reliance on fine-grained information from seen samples may lead to overfitting and damage generalization capability.

\section{Conclusion}
In this paper, we utilize visual concepts naturally embedded in local tokens of visual networks to enhance the text prompt with fine-grained information. Specifically, we introduce TextRefiner, a plug-and-play module, containing the local cache, feature aggregation, and feature
alignment. The local cache is used to continuously store fine-grained information from local tokens. Meanwhile, feature aggregation provides a solution to fuse global and local information to enhance the representation capability for the text branch. Feature alignment module can alleviate the modality gap between the text embedding and local tokens.
We also introduce two additional training losses, the semantic loss and regularization loss, to aid in the optimization process. TextRefiner can be seamlessly integrated into existing methods and introduce almost no additional computational overhead during inference. 
Our work offers new insights into improving the fine-grained representation capability of VLMs and efficient transfer learning in low-data scenarios. We hope this study will inspire additional advancements in the field of multimodal learning and efficient transfer learning.

\section{Acknowledgments}
This work was supported by National Science and Technology Major Project (No. 2022ZD0118201), the National Science Fund for Distinguished Young Scholars (No.62025603), 
the National Natural Science Foundation of China (No. U21B2037, No. U22B2051, No. U23A20383, No. U21A20472, No. 62176222, No. 62176223, No. 62176226, No. 62072386, No. 62072387, No. 62072389, No. 62002305 and No. 62272401), and the Natural Science Foundation of Fujian Province of China (No. 2021J06003, No.2022J06001).

\bibliography{aaai25}

\end{document}